# Anomaly Detection in General Ledger Data: Results from a Hybrid Approach

Jan Gronewald[1], Alexander Michael Rombach[1],
Sebastian Stephan[1], Peter Fettke[1]

[1]German Research Center for Artificial Intelligence (DFKI) GmbH
firstname.lastname@dfki.de

**ABSTRACT**

Journal Entry Tests (JETs) are a mandatory part of annual audits to evaluate and assess both high risk audit areas and potential material misstatements. However, as JETs are designed to detect known patterns based on domain knowledge, the resulting lists are often very large and require substantial additional effort from the auditor. To ensure the economic efficiency of the audit, the number of false positives in JET result lists must be reduced. Especially machine learning (ML) methods represent a promising approach to improve anomaly detection in this field. In this research in progress paper, we investigate different approaches on how to combine JETs with ML-methods in a hybrid manner. We present specialized models to increase the detection performance and validity of anomaly detection results to improve audit efficiency. The experiments are based on synthetic data consisting of different normal and anomalous journal entries.

## I. INTRODUCTION

The auditor's objective is to identify and assess the risks of material misstatement of financial statements due to fraud or errors. In accordance with ISA 240 and IDW PS 210, this includes planning and performing appropriate audit procedures to obtain reasonable assurance about whether the annual financial statements of the audited company are free from material misstatement, whether a misstatement was made intentionally or unintentionally (IAASB, 2013; Institut der Wirtschaftsprüfer (IDW), 2012). In this context, journal entry tests (JETs) are a mandatory component of the annual audit, in which the entries in a company's general ledger are audited. Due to the inherent complexity of journal entries, however, JETs primarily rely on known patterns, rules, and best practices (Droste & Tritschler, 2018). By this, they can flag potentially suspicious entries. However, JETs often lead to extensive result lists that auditors must sift through to identify true anomalies. The evaluation of these results entails longer processing times, a task that has become increasingly cumbersome over recent years. Motivated by the increase of data, clients also have not demonstrated a corresponding willingness to pay higher fees for audits. This development leads to an increasing pressure for auditing firms to enhance their operational efficiency while maintaining or improving the audit reliability.

To address these challenges, auditors are seeking ways to streamline the JET process. The objective is to achieve higher audit efficiency without compromising the quality and reliability of their tests. Enhancing efficiency might involve leveraging advanced data analytics tools, implementing more sophisticated machine learning (ML)-based anomaly detection methods, or refining the rules used in JET (e.g. finding new and unknown patterns). Research in this field has explored a variety of approaches to anomaly detection, both in financial statements and in general data. These studies aim to develop techniques that can more accurately identify unusual transactions, reducing the number of false positives (FPs) and enabling auditors to focus their efforts on genuinely suspicious entries. However, less attention is given to the combination of rule-based and ML-based methods so far.

Subsequently, in this research in progress paper we investigate how a hybrid anomaly detection approach can address mentioned challenges and improve audit efficiency. Inspired by integrating established rule-based methods and ML-based methods for anomaly detection, we research out the combination of both approaches to capitalize on their respective advantages. For this reason, we present two hybrid approaches. The first approach focuses on the identification of anomalies detected by JETs and ML-methods. By applying both methods independently, we want to identify potential anomalies by intersecting the result sets. The second approach entails deriving novel features from the JET result lists and incorporating them as supplementary features for ML-model training. For both approaches, we expect that the number of FP, but also false negatives (FNs) can be reduced. Evaluation takes place on a synthetic generated data set.

This paper unfolds as follows: After this introduction, section 2 discusses existing literature of JET and anomaly detection methods that are applied in the financial auditing domain. Section 3 presents our experimental design. We provide an overview of our business case study, our synthetic data generation approach, the anomaly detection pipeline and the evaluation metrics used. In section 4 we present our results, a discussion of the results, and limitations of our approach. Finally, section 5 concludes with a summary of the major results and an outline of future work.

## II.BACKGROUND

### Journal entry testing

The requirement to test journal entries arises from the International Standards on Auditing (ISA) issued by the International Auditing and Assurance Standards Board (IAASB) (IAASB, 2019). The occurrence of material misstatements should essentially be avoided by the existence of an effective internal control system (Marten et al., 2020). However, they have their limitations, which means that they cannot be designed to completely prevent material misstatements (Lanza &

Gilbert, 2007). For this reason, the auditor is required to analyze and evaluate financial reports and journal entries for the risk of material misstatement (IAASB, 2015, 2019).

To minimize the risk of material misstatement, the audit must be extended to the level of individual journal entries (IAASB, 2013). Therefore, there are different types of JETs discussed in literature (Center for Audit Quality, 2008; Droste & Tritschler, 2018; Kronfeld & Krenzin, 2014).

While all the mentioned approaches have in common to pinpoint unusual transactions that may indicate errors or fraudulent activities, they are limited due known patterns and the knowledge of the auditors. In addition, depending on selected filter criteria, JETs can lead to a high number of false positives. Here, more sophisticated anomaly detection methods provide a promising approach to address these limitations.

**Anomaly detection**

There is a plethora of anomaly detection methods discussed in literature. For example, Argyrou (2013) applies extreme value theory (EVT) to identify and analyze unusual or outlier journal entries. By leveraging EVT, the idea is to pay attention to transactions that deviate significantly from typical patterns. Baader & Krcmar (2018) present a methodology that integrates red flag analysis with process mining techniques to enhance the accuracy of fraud detection. By combining these approaches, they aim to minimize false positives and improve the identification of fraudulent activities within financial processes and thus journal entries. Guha & Gebremariam (2019) apply different ML-algorithms like K-means clustering and classification algorithms like random forest, SVM linear kernel, etc.) and evaluate them against each other to provide best practices to implement them, especially when dealing with imbalanced data. A different approach is pursued by Guo et al. (2022) who explore the use of graph topology to represent accounting data visually and leveraging the structural relationships between journal entries to identify fraudulent behavior.

Other approaches discussed in this field are for example (Al-Hashedi & Magalingam, 2021; Bay et al., 2006; Becirovic et al., 2020; Debreceny & Gray, 2010; Ngai et al., 2011; Seow et al., 2016; Wang, 2010).

Beside traditional data mining techniques, the use of ML and in particular deep learning, approaches like (deep) autoencoders open up new and promising possibilities. For example, Schreyer et al. (2017, 2019) implement a deep autoencoder to detect local and global anomalies in SAP-datasets. The work included a comparative evaluation against different ML-Algorithms like Principal Component Analysis (PCA), Hierarchical Density-Based Spatial Clustering of Applications with Noise (HDBSCAN), Local Outlier Factor (LOF) and One Class Support Vector Machines (OC-SVM) with superior detection results for one of the optimized Autoencoders. They further continued their work with the investigation of adversarial autoencoders to improve the interpretability of detected anomalies. Zupan et al. (2018) evaluate the applicability of variational autoencoders in journal entry data. In (Zupan et al., 2020), the authors extend their anomaly detection by a hybrid approach combining a variational autoencoder with a Long short-term memory (LSTM)-model. Another approach is presented by Schultz & Tropmann-Frick (2020), where they filter the journal entries by accounts and trained autoencoders on each subset. They have shown that the detection of an autoencoder still works on prefiltered journal entries, and thus small data sets. Bakumenko & Elragal (2022) present another approach where they evaluated besides several supervised approaches a Deep Autoencoder and Isolation Forest. Using synthetic anomalies, the evaluation has shown that the number of false negatives and false positives could be reduced.

As shown, many anomaly detection approaches are discussed in literature However, to the best of our knowledge, a hybrid approach combining JET and ML-methods like autoencoders has not been considered so far. For this reason, we investigate how the combination can leverage the advantages of both approaches to reduce false positives.

## III.EXPERIMENTAL DESIGN

### Overview

In the following sections, we present our experimental design (Figure 1). First, we provide general information regarding the posting context and the business cases used to generate synthetic general ledger data (journal entries). We then describe our synthetic data generation pipeline, the standard business cases we have considered and how postings, but also synthetic anomalies (anormal postings) were generated. In addition to the applied JETs, we describe the data preprocessing procedures for applying different ML-methods like Isolation Forest, HDBSCAN, and a Deep Autoencoder. This encompasses how we preprocessed the data in terms of their multi-line posting structure, as well as the processing of categorical and numerical features. Finally, we describe the implementation of anomaly detection methods. We commence with the application of JETs and ML-methods before progressing to the application of our hybrid anomaly detection approaches 1 (H1) and 2 (H2). In H1, we apply JETs and ML-methods independently and use the intersection of the identified anomalies as final result. In H2 we use the JET result lists as input to consider JET-specific features within the ML-model to improve model performance.

### Business Case Study

The case study describes a trading company for garden tools. For the sake of simplicity, we focused only on following four business cases: (1) purchase of merchandise, (2) sale of goods, (3) payments (incoming and outgoing), and (4) various expenses. In (4), postings such as monthly recurring expenses from payroll, expenses from various leasing agreements, service contracts, and rents and other expenses for the operation of the business premises, are included. In sum, there are six accountants, who are responsible for the posting of the described business cases. The responsibility and frequency of who posts which business transactions is not uniform at all and is determined randomly. As account chart, we used the standard DATEV SKR03 (DATEV eG, 2024).

Outgoing of the business cases, we have selected four dedicated JETs. The selection process was designed to ensure that the resulting data set closely aligns with the characteristics of the simulated business transactions. The definition of the JETs is based on (Droste & Tritschler, 2018). Based on expert interviews, we then considered following four JETs: (J1) top n highest postings, (J2) high disbursements from cash accounts, (J3) timely posting (promptly), and (J4) postings at unusual times (e.g. outside official working hours) and days (e.g. weekend).

Assuming that postings are only made during regular working hours and that only the six accountants are allowed to post business transactions, we have simulated regular postings, which can, but don’t have to fall in the range of our JET patterns.

**Generation of synthetic data and anomalies**

There are many applications and methods discussed in literature how to generate synthetic data (Emam et al., 2020). As we did not have any real posting data, we decided to generate synthetic data. Our approach is based on the general accounting logic. The definition and simulation of postings is based on the author’s knowledge in this area.

According to our case study, we focused on the generation of dedicated business transactions following the DATEV posting structure. Thus, a multi-line representation was selected to define individual posting records. Related posting records are identified by their *posting_id*. In sum, we consider three timestamp attributes, namely *document_date*, *posting_date*, and *entered_date*. These were considered to carry out various evaluations, such as checking if postings be booked promptly or on an accrual basis. Other important attributes such as the account (*gl_account*), the account name (*gl_account_name*) and the *amount* were also recorded. How the account in question is posted is described by the debit/credit flag (*cd_flag*) of the respective line item. In addition to the *user* creating the posting, the tax rate (*tax_rate*) is specified. A more detailed description of each data attribute is given in Table 1.

For generating account postings, we have made different assumptions. First, we generated journal entries only for one year. Second, to differentiate between items which are purchased, paid, and be resold, we defined a static price list for each item. The purchase price was derived by the material costs of each item. The selling price is calculated based on the purchase price by multiplying it with a static factor. Third, time intervals and the number of postings are randomly set based on different probability distributions like discrete uniform distribution, uniform distribution, normal distribution, and Poisson distribution.

The calculation consists of following steps: First, a list of dates within a time period was generated. In our use case, we choose 01.01.2023 – 31.12.2023. For each date, we then generated multiple timestamps. For weekdays, we have chosen a Poisson distribution with lambda=15. We used this to model the number of events per date by assuming that 15 booking transactions are carried out on average. For the number of timestamps on a weekend day (Saturday or Sunday), we used a normal distribution with a mean of 10 and standard deviation of 2. To provide timestamps for entry date, we defined normal working hours (8 AM – 5 PM) and non-working hours (< 8 AM or > 5 PM). Based on an individual probability, we then generate posting transactions within regular working hours. Timestamps outside the regular working hours before 8 AM or after 5 PM were selected by a 50% probability each. Defining weights for weekdays and months respectively, we were able to weight certain days or month. To do so, the weights are normalized to create a probability distribution, which was then used to randomly select an entry date from a set of timestamps within the defined time period and day. To model time differences between an entry date and a document or posting date, we used the truncated normal distribution. To simulate a normal deviation, we used an interval of [0;10] with a mean of 5 and standard deviation of 2. Based on a probability threshold, we were able to simulate higher time intervals using the same distribution within an interval of [10;60]. Finally, the time deltas were subtracted from the *entered_date* to derive the document and posting date.

In Table 2, an overview of the generated postings is given. As shown, different postings with different parameter settings based on our predefined business cases are defined. For example, in case 2, purchase postings were generated where Thursdays and Fridays have a slightly higher chance to be selected. By this, we tried to consider different posting behavior. Users are selected randomly for each posting. Another example is shown in case 3 where we generated postings where only one items was sold (single), but also multiple items of the same type (multiple). In case 4, for example, we generated only transactions within a specific time period. Using different weights, we simulate that more postings are made in the summer months. Finally, as shown in case 9, we set a time delta probability of 80%, which means that we want to add with a probability of 20% a time delta > 10 days. By this, we were able to consider postings which are, for example, less promptly. In sum, we generated 51,076 business transactions (127,835 posting lines)[1]. All transactions consist of either two or three posting lines.

Next, for the generation of synthetic anomalies, we focused on attribute combinations which are not existent in the journal entry data. Synthetic anomalies are generated based on our four journal entry tests. Synthetic anomalies were inserted based on non-existent combinations in our generated journal entries, which are labeled as normal and not existent in the relating JET result list. For example, 20 transactions were randomly selected for J1 within the 1000 highest amounts and then the account and user values were changed. For J2, transactions with an amount > 1,500 euros were considered. Afterwards, 10 random transactions were selected, and the user was changed, too. According to the time thresholds in J3, only anomalies that were posted less promptly or not at all promptly were included. To do so, we used time deltas between 0-9 days to classify postings as promptly, whereas time deltas between 10-29 and 30+ are considered as less promptly or not at all promptly respectively. Finally, based on J4, the booking transactions that were strongly underrepresented depending on the user (e.g. < 10 postings) were marked as anomalous. We made sure that synthetic anomalies made sense in terms of content, for example, in the event of posting other vehicle costs, which is usually booked during business hours, was then suddenly booked outside of working hours. Finally, weekend postings were added where the user was not represented in the results list and an unusual account was posted to. The changed posting records were marked accordingly so that they could be used as (labeled) anomalies for evaluation purposes. During generation, we tried to ensure that previously generated anomalies were not changed. In total, we generated 573 anomalies.

[1] Postings, which belong to 2022 or 2024 respectively were filtered out to focus on one closed year. However, we still considered for example purchase transactions which have no relating payments.

**Data preprocessing**

The data was preprocessed for subsequent application of the ML algorithms. Therefore, certain fields were excluded from further consideration. This included the *document_date*, *posting_date*, *entered_date*, *gl_account*, *ref_id*, *text* and the *posting_id* attributes. The *posting_date* has been removed due to its redundancy with the *document_date*. Furthermore, the difference between the attributes *document_date* and *entered_date* was calculated to be able to consider documents that are not posted promptly. The difference was included in the data set as an additional attribute (feature) and the actual date attributes were excluded from the further process. In addition, the *gl_account* attribute has been deleted as it contains redundant information due to the correlation with the *gl_account_name* attribute. The *ref_id* and *text* attributes were not considered in the ML-model and were therefore deleted. Since the *posting_id* attribute only groups the individual posting lines, it was dropped, too. It was not required for the ML-models (model) as the postings were merged at document level.

Since the models we used cannot process categorical values directly, we preprocessed categorical attributes using a one-hot encoding. This affected the attributes *gl_account_name*, *dc_flag*, *user*, and *tax_rate*. The numerical attributes with the *date_difference* and *amount* were first scaled using a so called RobustScaler[2] to reduce the impact of outliers. Subsequently, both fields were scaled to the interval [0;1] using MinMaxScaler[3] like described in (Bakumenko & Elragal, 2022).

To preserve the relation between the entries, they were merged at document level. Following the approach described in (Bakumenko & Elragal, 2022), we merged all transactions belonging to the same posting by concatenating their feature vectors. This resulted in 51,076 rows, where each row represents one journal entry. Since journal entries can consist of different numbers of transactions, we padded all rows to the maximum length – in our case 3 transactions. “Empty” cells of category features were padded with a “PAD” category, while the amount columns were padded with 0-values. Due to the unsupervised learning characteristics of our three ML-models, the attribute *label* was separated from the data set. At the end of the pre-processing pipeline, we obtained a feature vector of length 92, which served as input for our ML-baseline-models. For H1, an indicator for success (0) or failure (1) was created for each JET. In H2, the aforementioned features were also encoded one-hot and included in the model training. This led to increased feature vectors with a length of 104.

---

[2] https://scikit-learn.org/stable/modules/generated/sklearn.preprocessing.RobustScaler.html, accessed on: 07.07.2024
[3] https://scikit-learn.org/stable/modules/generated/sklearn.preprocessing.MinMaxScaler.html, accessed on: 07.07.2024

## Anomaly detection pipeline

### ***Journal Entry Tests***

To identify and review the largest journal entries, we applied J1 to get the postings that can have a significant impact on the financial statements. In our experiment, we used a threshold of n=1,000. While cash is highly susceptible to fraud, focusing on high disbursements from cash accounts, J2 is also of interest to scrutinize these transactions to confirm they are legitimate. However, the challenge relies on setting an appropriate threshold. For our use case, we set the amount > 1,500 euros. J3 deals with timely postings. Timely posting is crucial for maintaining accurate financial records and that financial statements are reliable. According to Droste & Tritschler (2018), different time intervals are used to limit the posting records. Thus, we set three thresholds to check for timely posting: If the difference between the attributes *document_date* and *entered_date* is between 0-9 days, we consider such transactions as promptly and thus not critical. However, if the time delta is greater than > 10 days, we treat them as not promptly.

Finally, we considered postings outside regular working hours and on weekend days. Similar to the other JETs, journal entries within the result list do not automatically constitute fraud or an anomaly. However, it can be useful to analyze in more detail when postings were made. By taking other attributes such as *amount*, gl*_account* / *gl_account_name*, *cd_flag* into account, a targeted selection can be made that enables an assessment of whether these are plausible or not.

However, all the mentioned JETs have in common that their result lists can consist of many posting records, which have to be checked by the auditor. Thus, according to our expert interviews, new approaches are needed to reduce FPs and FNs to improve operational efficiency. Due to its relevance, however, the challenge in the application of JETs relies on setting a proper threshold in order to get manageable result lists with less FPs and FNs.

***Machine learning Approaches***

Given the fact that in real-world audit settings data is usually not labelled and that we are not aiming for a generally valid model with our approach, but rather training for each audit mandate, we decided to use exclusively unsupervised learning models. In particular, we opted for a Deep Autoencoder, Isolation Forest, and HDBSCAN. In the following, we present the models and describe the hyperparameters used.

Deep Autoencoders are a type of neural network architecture, which is designed to compress the journal entries into a lower dimensional latent representation and to reconstruct them from the latent representation back into the output layer. We chose a network architecture with one input layer, seven fully connected hidden layers and one output layer. The size of the input layer depends on the length of the feature vector. This differed between the baseline model (length: 92), which was also used in H1, and H2 (length: 104). This resulted in the following layer sizes: [92, 36, 18, 9, 6, 3, 6, 9, 18, 36, 92] for the baseline and H1 and [104, 36, 18, 9, 6, 3, 6, 9, 18, 36, 104] for H2. Each of the layers was implemented as a linear layer using a Sigmoid Linear Unit (*SiLU)*[4] activation function between each layer. For the model training, we used a batch-size of 128 with a learning rate of 10^-4 for 200 epochs with early stopping. Further, we used the *Mean Squared Error (MSE) loss*[5] and *Adam optimizer*[6]. The *MSE loss* function was used because it is suitable for measuring reconstruction errors (e.g. penalizing larger errors more than smaller ones), while the

[4] https://pytorch.org/docs/stable/generated/torch.nn.SiLU.html, accessed on: 07.07.2024
[5] https://pytorch.org/docs/stable/generated/torch.nn.MSELoss.html, accessed on: 07.07.2024
[6] https://pytorch.org/docs/stable/generated/torch.optim.Adam.html, accessed on: 07.07.2024

*Adam optimizer* provides a robust and efficient optimization method, facilitating the training of the autoencoder. The models were implemented in *PyTorch*[7].

HDBSCAN is a hierarchical clustering algorithm that extends the Density-Based Spatial Clustering of Applications with Noise (DBSCAN) algorithm by considering a hierarchy between clusters. It searches the data space for high-density regions and combines them into clusters, then merges smaller clusters to extract stable clusters. It can be used to perform anomaly detection by locating data points outside of high-density regions and subsequently flagging them as anomalies. The algorithm can be run with a minimum number of hyperparameters. In our case, we chose to set the minimum cluster size to just 12 to allow the algorithm to cluster patterns of postings that occur only once a month.

Isolation Forest is a tree-based algorithm that recursively divides the data set into random partitions until each data point is isolated from the other data points. During this process, anomalies are isolated from the remaining data points faster than normal data points. Accordingly, the path length is documented during the division and converted into an anomaly score per data point. The Isolation Forest can be run with a minimum number of hyperparameters. We chose to set the contamination parameter to *auto* in order to avoid adding any expectations in the model. In addition, the number of trees to be trained was set at 650 to ensure sufficient fidelity.

***Hybrid Approaches***

As part of our research, we tested two different hybrid approaches combining ML methods and JET. Our motivation results from the idea to leverage the strengths of different approaches and thus to increases the robustness of the anomaly detection (Chandola et al., 2009; Hodge & Austin, 2004). In addition, anomalies can manifest in various forms making it necessary to look from different perspectives on the data. For example, Schreyer et al. (2017) distinguish between local and global anomalies highlighting the challenges in identifying these different types of anomalies. While rule-based systems are effective at identifying known issues based on established rules, ML can uncover previously unknown patterns and correlations (Bhattacharyya et al., 2011; Gepp et al., 2018; Perols, 2011; Perols et al., 2016). Thus, using different anomaly detection approaches, we are convinced that a dual approach reduces the likelihood of FP and enhances the overall performance of the audit process.

[7] https://pytorch.org/, accessed on: 07.07.2024

Our first hybrid approach (H1) deals with the independent application of JETs and ML-models. By intersection the result lists, we focus our analysis only on those postings, which are recognized by both approaches. H1 is based on the consideration that by intersecting the result lists, the FPs can be reduced. A different approach is followed in approach H2. Here, JETs are executed first. The results and information obtained are then considered as additional features in the data and used as additional features within model training. For example, checking for promptly postings, we labeled each posting according to the thresholds used (J3). Here, we follow the idea, not to intersect only the result lists. Rather, we follow the idea to provide additional information to the model, which can have different meanings depending on the company and audit context to increase TP. Both hybrid approaches were tested against our ML- and JET baseline models. While the models were trained on the whole data set, the JET results were summarized.

**Evaluation metrics**

For evaluation purposes, we use precision, recall, F1 and ROC-AUC scores[8] as defined in (Sokolova & Lapalme, 2009). These metrics are widely used in the field of anomaly detection and classification problems helping us to draw conclusions about the TPs and FPs. For example, a low

[8] https://scikit-learn.org/stable/modules/model_evaluation.html#classification-report, accessed on: 07.07.2024

precision would mean that the model still identifies TPs, but also makes a lot of FPs. They are based on the calculated results from the confusion matrix, which is provided in Table 3. Since the underlying data is highly imbalanced (in terms of normal/anormal postings), we consider the macro average[9] of the metrics, which weight the prediction performance regarding both classes equally (Sokolova & Lapalme, 2009).

In our case, we formulate the anomaly detection problem as a binary classification task where the entries must be classified as either normal or anomalous. Since the anomaly detection of our Deep Autoencoder is based on a reconstruction error and not on a binary classification, we have defined a threshold for this model that separates the data into a partition with normal data and a partition with anomalous data to obtain predictions (and thus labels). Similar to Bakumenko & Elragal (2022), we classified a data point as anomalous if its reconstruction error exceeded the 99th percentile of the errors observed in the entire dataset. The other two models allowed a binary classification by default.

## IV.EVALUATION

### Results

Table 3 shows the results of our experiments. First, we implemented the JETs. For the sake of simplicity, we did not distinguish between each JET result list. The JET-baseline includes all 573 synthetic anomalies and thus identified them correctly (TPs). However, this was expected because we have injected anomalies only inside the JET result lists. The JET result list, however, contains in sum 10,019 FPs, which must be sifted by the auditor. This illustrates the problem mentioned at the beginning quite well. While no other anomalies as described were injected, we have 0 FNs and therefore 40,484 TNs.

---

[9] https://scikit-learn.org/stable/modules/generated/sklearn.metrics.classification_report.html, accessed on: 07.07.2024

Next, we implemented our three ML-models and evaluated them along our aforementioned evaluation metrics. The predictions were used as baseline for further comparisons, especially with regard to H1 and H2. As shown in Table 3, none of the ML-models was able to detect all synthetic anomalies. The Deep Autoencoder achieved acceptable results in terms of precision and F1 score. However, we also see that significantly more FNs are predicted in relation to the TPs. A high number of FNs may suggest that the detection method is not efficiently covering the entire journal entries or is not well-tuned to capture all types of anomalies. Due to the model complexity, a hyperparameter optimization regarding network architecture, learning rate, regularization etc. might be useful. In contrast to the Deep Autoencoder, the Isolation Forest achieved the best results in terms of recall and ROC-AUC score. Detecting 554 TPs and only 19 FNs, it indicates promising results. Obviously, the model can better distinguish between transactions that are considered normal and abnormal. Despite a high recall, however, the model leads to a low precision, which means that the model identifies many TPs, but also makes a lot of FPs. One reason might be due the generated data. While the model detects outliers, obviously the performance degrades if the data contains a high number of noise points. Finally, HDBSCAN lagged behind the other two ML-models, providing average results between 51-53 %. The reason for that might be the usage of default parameter setting due HDBSCAN heavily relies on its configuration regarding cluster size, number of samples, distance metrics, and cluster selection methods.

In approach H1 there was no performance improvement with respect to the identified TPs and no decrease in classifying FNs. However, the number of FPs was reduced. The Deep Autoencoder achieved the best result in terms of precision, and the Isolation Forest achieved the best results in terms of recall, F1 and ROC-AUC score. In approach H2, the classification with regard to the detection of TPs could be increased in all models and thus improved compared to the plain ML-baseline and H1. In contrast to the other two algorithms, all 573 anomalies were detected by the Isolation Forest. Accordingly, it still performed best in terms of recall and ROC-AUC score, while the Autoencoder performed almost unchanged best in terms of precision and F1 score. However, it is interesting to see that providing JET-related information, the performance can be slightly improved.

**Discussion**

In the context of auditing, it is particularly important that no material errors are overlooked. For this reason, the objective is to identify as many TPs as possible, but also to decrease FPs in order to improve audit efficiency. Against this background, we evaluate the model quality primarily regarding precision and recall. Regarding our baseline models, a reduction in FPs was achieved for all models.

This result was significantly improved by H1 in all algorithms, but still at the cost of 19 undetected anomalies. This shows that the simple combination of JET and ML already brings considerable efficiency advantages. However, the delta of the result list should still be checked for material misstatements as these can still contain high value errors.

From our perspective, the highest quality of results was achieved in variant H2. The models trained there show the best performance in terms of recall in comparison between our three baseline-models. The Isolation Forest was able to achieve complete detection of all anomalies while at the same time significantly reducing the FPs from 10,019 to 1,881 entries. The H2 variant contains 667 more FPs than the H1 variant, but in return has complete detection of all anomalies. Interestingly, HDBSCAN does not perform very well in all setups. It performs poorly on the metrics we measured and has limited parameters for further optimization. Slightly better results were achieved with the Deep Autoencoder. This applies in particular to the ratio of TP to FP. The Deep Autoencoder achieved the best performance in this area but was far from being able to detect all anomalies. Nevertheless, we see further potential for in-depth testing of the model in our application context thanks to the various optimization and architecture options.

In connection with H2 and the IF, we also asked ourselves what influence the remaining FPs have on the performance of the approach and whether there is any connection with the synthetically generated data. After a more detailed descriptive analysis, we were able to determine that at least one of the JETs scored a hit in 1,571 of the total 1,881 entries, which partially explains the labeling as an anomaly. Of the remaining 310 entries, 284 entries can be assigned to the purchase and sale of a specific item. The remaining 26 entries are distributed relatively heterogeneously across different cases. Here a precise explanation of the anomalies would require an in-depth analysis.

**Limitations**

The implemented anomaly detection approach exhibits several limitations, primarily due to the constraints and assumptions made. The use of synthetic data, generated as a substitute for real-world data, introduces fundamental challenges. It was produced based on less complex assumptions, reflecting only a well-defined real-world posting context. In addition, the generated postings are characterized by randomly generated values. The probability distributions could have been more finely tuned to reflect the specific business cases under consideration or to simulate different posting behavior. Furthermore, the variability of generated values regarding the accounts, user, and amounts could be enhanced to reflect more accurately the posting complexities. In this way, more demanding booking cases, such as those relating to discounts, goods complaints and ancillary bookings, could be considered, which would give the data record the necessary depth. Furthermore, it also remains unclear to what extent it is possible to detect anomalous journals, as there are various patterns and anomalies where different transaction types may exhibit unique behavior. Also, fraudulent entries are often designed to be unnoticeable or to resemble regular booking records. Accordingly, it is necessary to consider the context in which transactions take place.

Also, the ML-models are based on default parameter settings and standard preprocessing steps such as one-hot encoding and the concatenation of related postings. While functional, these methods could benefit from the integration of more advanced encoding techniques, such as entity embeddings (C. Guo & Berkhahn, 2016) or large language model embeddings (Bakumenko et al., 2024), which would likely yield more nuanced and reliable results.

Finally, the applicability of the presented approaches on a real-world journal entry data or within an audit setting remains a critical concern. The synthetic nature of the data and the presented anomaly detection methods may not fully capture the intricacies and variabilities of actual financial transactions. Consequently, the effectiveness and reliability of these approaches in real-world scenarios remain uncertain, underscoring the need for further refinement and validation using more nuanced and comprehensive synthetic and genuine data.

## V.CONCLUSION AND FUTURE WORK

In this research in progress paper, we present an approach how hybrid anomaly detection approaches can help to reduce the number of false positives in JET result lists. By using synthetic data, the application of the presented hybrid approaches was shown. Compared to standard implementations using dedicated JETs and ML-methods, we have shown that the detection of anomalies can be slightly improved. However, the limitations are still manifold.

Guided by the presented results, our future research consists of various tasks. First, we will enhance our synthetic data generation pipeline. Synthetic data serves as a foundational element for researching anomaly detection in account postings. By improving this pipeline, we aim to create high-quality synthetic data sets, not only for the sake replicability, but also to be used to develop new anomaly detection approaches in this field. This also applies to the possibility of taking targeted account of anomalous booking data and patterns. However, due to the limitations of relying solely on synthetic data sets, we also plan to conduct evaluations on real-world data. Evaluation on real-world data is crucial to validate their effectiveness and applicability in practical scenarios, but also to determine how reliable our synthetic data generation pipeline is. Moreover, we plan to extend our research to consider additional JETs and ML-methods. By comparing a wider range of methods and techniques, we aim to present a more comprehensive picture of the possibilities, best practices, but also limitations of existing approaches. In combination with synthetic data, we will focus on how the performance of ML-methods or hybrid approaches can be improved. Finally, we will implement another hybrid approach where JETs are used as prefilters, and ML-models are trained on the result lists. The objective of this method is to recognize complex patterns and relationships within these result lists to reduce FPs and increasing TP.

**REFERENCES**

Al-Hashedi, K. G., & Magalingam, P. (2021). Financial fraud detection applying data mining techniques: A comprehensive review from 2009 to 2019. *Computer Science Review*, *40*(C). https://doi.org/10.1016/j.cosrev.2021.100402

Argyrou, A. (2013). Auditing journal entries using extreme value theory. *ECIS 2013*.

Baader, G., & Krcmar, H. (2018). Reducing false positives in fraud detection: Combining the red flag approach with process mining. *International Journal of Accounting Information Systems*, *31*, 1–16. https://doi.org/10.1016/j.accinf.2018.03.004

Bakumenko, A., & Elragal, A. (2022). Detecting Anomalies in Financial Data Using Machine Learning Algorithms. *Systems*, *10*(5). https://doi.org/10.3390/systems10050130

Bakumenko, A., Hlaváčková-Schindler, K., Plant, C., & Hubig, N. C. (2024). *Advancing Anomaly Detection: Non-Semantic Financial Data Encoding with LLMs* (ArXiv).

Bay, S., Kumaraswamy, K., Anderle, M. G., Kumar, R., & Steier, D. M. (2006). Large scale detection of irregularities in accounting data. *Sixth International Conference on Data Mining (ICDM06)*, 75–86. https://doi.org/10.1109/ICDM.2006.93

Becirovic, S., Zunic, E., & Donko, D. (2020). A Case Study of Cluster-based and Histogram-based Multivariate Anomaly Detection Approach in General Ledgers. *19th International Symposium INFOTEH-JAHORINA, INFOTEH 2020*. https://doi.org/10.1109/INFOTEH48170.2020.9066333

Bhattacharyya, S., Jha, S., Tharakunnel, K., & Westland, J. C. (2011). Data mining for credit card fraud: A comparative study. *Decision Support Systems*, *50*(3), 602–613. https://doi.org/10.1016/j.dss.2010.08.008

Center for Audit Quality. (2008). *Practice Aid for Testing Journal Entries and Other Adjustments Pursuant to AU Section 316*. https://us.aicpa.org/content/dam/aicpa/interestareas/centerforauditquality/resources/caqauditlibrary/downloadabledocuments/caq-practice-aid-for-testing-journal-entries.pdf

Chandola, V., Banerjee, A., & Kumar, V. (2009). Anomaly detection: A survey. *ACM Computing Surveys*, *41*(3). https://doi.org/10.1145/1541880.1541882

DATEV eG. (2024). *Account Chart. Standard Chart of Accounts (SKR 03)*. https://www.datev.de/web/de/datev-shop/material/skr-03-englisch/

Debreceny, R. S., & Gray, G. L. (2010). Data mining journal entries for fraud detection: An exploratory study. *International Journal of Accounting Information Systems*, *11*(3), 157–181. https://doi.org/10.1016/j.accinf.2010.08.001

Droste, K. C., & Tritschler, J. (2018). *Journal Entry Testing*. IDW Verlag.

Emam, K., Mosquera, L., Hoptroff, R., & Safari, an O. M. Company. (2020). *Practical Synthetic Data Generation* (1st ed.). O'Reilly.

Gepp, A., Linnenluecke, M. K., & Smith, T. (2018). Big Data in Accounting and Finance: A Review of Influential Publications and a Research Agenda. *Journal of Accounting Literature*, *July 2017*. https://www.researchgate.net/publication/316616708

Guha, A., & Gebremariam, E. (2019). An Exploratory look at Data Extraction and Machine Learning for Detecting Fraudulent Financial Journal Entries. *International Journal of Engineering Research And*, *8*(9). https://doi.org/10.17577/ijertv8is090049

Guo, C., & Berkhahn, F. (2016). *Entity Embeddings of Categorical Variables* (ArXiv). http://arxiv.org/abs/1604.06737

Guo, K. H., Yu, X., & Wilkin, C. (2022). A Picture is Worth a Thousand Journal Entries: Accounting Graph Topology for Auditing and Fraud Detection. *Journal of Information Systems*, *36*(2), 53–81. https://doi.org/10.2308/ISYS-2021-003

Hodge, V., & Austin, J. (2004). A Survey of Outlier Detection Methodologies. *Artificial Intelligence Review*, *22*, 85–126.

IAASB. (2013). *INTERNATIONAL STANDARD ON AUDITING 240 THE AUDITOR'S RESPONSIBILITIES RELATING TO FRAUD IN AN AUDIT OF FINANCIAL STATEMENTS*. https://www.ifac.org/_flysystem/azure-private/publications/files/A012 2013 IAASB Handbook ISA 240.pdf

IAASB. (2015). *INTERNATIONAL STANDARD ON AUDITING 330 THE AUDITOR'S RESPONSES TO ASSESSED RISKS*. https://www.ibr-ire.be/docs/default-source/nl/Documents/regelgeving-en-publicaties/rechtsleer/normen-en-aanbevelingen/ISA-s/clarified-ISA-s/ISA-update-2015/English/A019-ISA-330-for-Handbook_formatted.pdf

IAASB. (2019). *INTERNATIONAL STANDARD ON AUDITING 315 (REVISED 2019) IDENTIFYING AND ASSESSING THE RISKS OF MATERIAL MISSTATEMENT*. https://www.ifac.org/_flysystem/azure-private/publications/files/ISA-315-Full-Standard-and-Conforming-Amendments-2019-.pdf

Institut der Wirtschaftsprüfer (IDW). (2012). *IDW Prüfungsstandard: Zur Aufdeckung von Unregelmäßigkeiten im Rahmen der Abschlussprüfung (IDW PS 210)*.

Kronfeld, T., & Krenzin, A. (2014). Analytische Forensic-Accounting-Verfahren zur Aufdeckung von Unregelmäßigkeiten in der Buchführung. *Betriebswirtschaftliche Forschung Und Praxis*, *2*.

Lanza, R. B., & Gilbert, S. (2007). A Risk-Based Approach to Journal Entry Testing. *Journal of Accountancy*, *204*(1), 32–35.

Marten, K.-U., Quick, R., & Ruhnke, K. (2020). *Wirtschaftsprüfung. Grundlagen des betriebswirtschaftlichen Prüfungswesens nach nationalen und internationalen Normen* (6th ed.). Schäffer-Poeschel.

Ngai, E. W. T., Hu, Y., Wong, Y. H., Chen, Y., & Sun, X. (2011). The application of data mining techniques in financial fraud detection: A classification framework and an academic review of literature. *Decision Support Systems*, *50*(3), 559–569. https://doi.org/10.1016/j.dss.2010.08.006

Perols, J. (2011). Financial statement fraud detection: An analysis of statistical and machine learning algorithms. *Auditing*, *30*(2), 19–50. https://doi.org/10.2308/ajpt-50009

Perols, J., Bowen, R. M., Zimmermann, C., & Samba, B. (2016). Finding Needles in a Haystack: Using Data Analytics to Improve Fraud Prediction. In *SSRN*. https://doi.org/10.2139/ssrn.2590588

Schreyer, M., Sattarov, T., Borth, D., Dengel, A., & Reimer, B. (2017). Detection of Anomalies in Large Scale Accounting Data using Deep Autoencoder Networks. *ArXiv*, *Aug*. http://arxiv.org/abs/1709.05254

Schreyer, M., Sattarov, T., Schulze, C., Reimer, B., & Borth, D. (2019). Detection of Accounting Anomalies in the Latent Space using Adversarial Autoencoder Neural Networks. *ArXiv*, *Aug*. http://arxiv.org/abs/1908.00734

Schultz, M., & Tropmann-Frick, M. (2020). Autoencoder neural networks versus external auditors: Detecting unusual journal entries in financial statement audits. *Proceedings of the*

*53rd Hawaii International Conference on System Sciences*, 5421–5430. https://doi.org/10.24251/hicss.2020.666

Seow, P. S., Pan, G., Suwardy, T., Seow, S. ;, Pan, G. ;, & Seow, P.-S. (2016). Data Mining Journal Entries for Fraud Detection: A Replication of Debreceny and Gray's (2010) Techniques. *Journal of Forensic and Investigative Accounting*, *8*(3), 501–514. https://ink.library.smu.edu.sg/soa_research/1515

Sokolova, M., & Lapalme, G. (2009). A systematic analysis of performance measures for classification tasks. *Information Processing & Management*, *45*(4), 427–437. https://doi.org/10.1016/j.ipm.2009.03.002

Wang, S. (2010). A comprehensive survey of data mining-based accounting-fraud detection research. *2010 International Conference on Intelligent Computation Technology and Automation, ICICTA 2010*, *1*, 50–53. https://doi.org/10.1109/ICICTA.2010.831

Zupan, M., Letinić, S., & Budimir, V. (2018). JOURNAL ENTRIES WITH DEEP LEARNING MODEL. *International Journal of Advanced Computational Engineering and Networking*, *6*(1), 55–58. https://www.iraj.in/journal/journal_file/journal_pdf/3-507-154545867755-58.pdf

Zupan, M., Letinic, S., & Budimir, V. (2020). Accounting Journal Reconstruction with Variational Autoencoders and Long Short-term Memory Architecture. *CEUR Workshop Proceedings*, *2646*, 88–99.

## NUMBERED FIGURES

FIGURE 1: Overview of our experimental design

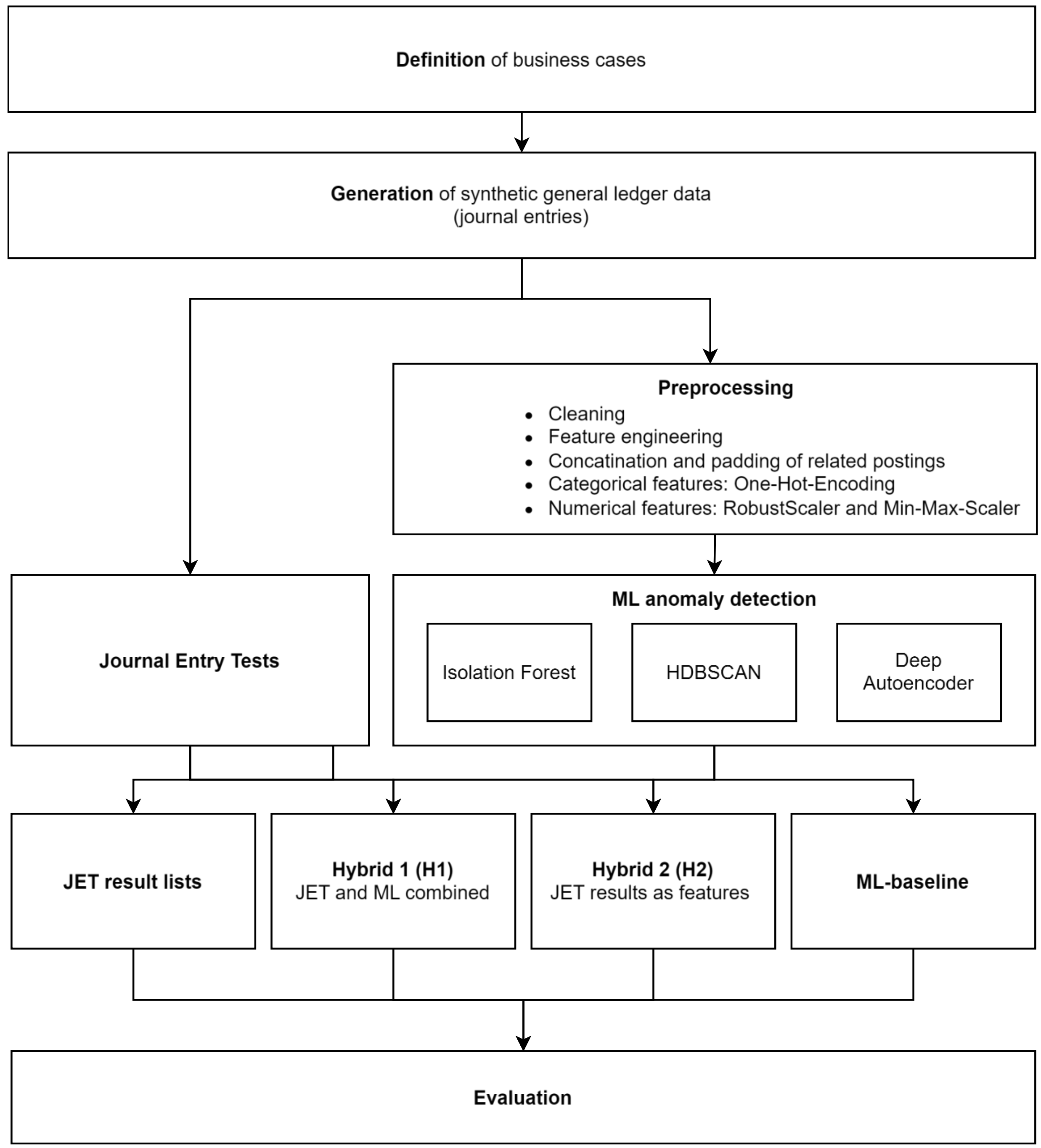

# NUMBERED TABLES

## TABLE 1

### Data attributes

| Name | Meaning | Datatype |
|---|---|---|
| posting_id | Unique ID to group account postings together. | alphanumeric |
| document_date | Manually set by user. Corresponds to the date of the document to be posted (e.g., invoice date). | date |
| posting_date | Manually set by user and is used to assign the account posting to the posting period (service date). | date |
| entered_date | Actual date and time when the document was entered by a user. Assigned by the system. | timestamp |
| gl_account | Account based on DATEV SKR03. | number |
| gl_acount_name | Name of the account used. | alphanumeric |
| amount | Net amount. | number |
| dc_flag | Debit-/credit-flag. Used to determine on which side the “gl_account” is booked. | text |
| user | Name of the user. | text |
| tax_rate | Applied tax rate on the amount. | number |
| ref_id | Used to reference to the posting which it clears. | alphanumeric |
| text | Posting text. | alphanumeric |
| label | Used to mark artificial inserted anomalies. | number |

**TABLE 2**

**Overview of generated bookings**

| Case | Name | Description | Amount |
|---|---|---|---|
| 1 | Purchase merchandise | Accounts: 3400, 1576, 1600<br>Days: Monday - Friday<br>Months: January – December 2023<br>Users: Alice, Bob, David<br>Items: multiple | 5,000 |
| 1 | Payment | Accounts: 1600, 1100<br>rest same as purchase merchandise | 5,000 |
| 2 | Purchase merchandise | Accounts: 3400, 1576, 1600<br>Days: Monday – Wednesday, Thursday (1,5), Friday (1,5)<br>Months: January – December 2023<br>Users: Alice, Bob, David<br>Items: multiple | 4,000 |
| 2 | Payment | Accounts: 1600, 1000<br>rest same as purchase merchandise | 4,000 |
| 3 | Sales of goods | Accounts: 1400, 8400, 1776<br>Days: Monday - Friday<br>Months: January – December 2023<br>Users: Alice, Bob, David<br>Items: single, multiple | 2,000;<br>1,500 |
| 3 | Incoming payment | Accounts: 1100, 1400<br>rest same as sales of goods | 2,000;<br>1,500 |
| 4 | Purchase merchandise | Accounts: 3400, 1576, 1600<br>Days: Monday – Friday<br>Months: March (1.1), April (1.2), May (1.3), June (1.5), July (1.7), August (1.7), September (1.3), October (1.2)<br>Users: Alice, Bob, David<br>Items: multiple | 4,000 |
| 4 | Payment | Accounts: 1600, 1100<br>rest same as purchase merchandise | 4,000 |
| 5 | Sales of goods | Accounts: 3400, 1576, 1600<br>Days: Monday – Friday<br>Months: March (1.1), April (1.2), May (1.3), June (1.5), July (1.7), August (1.7), September (1.3), October (1.2)<br>Users: Alice, Bob, David<br>Items: multiple | 4,000 |
| 5 | Incoming payment | Accounts: 1100, 1400<br>rest same as sales of goods | 4,000 |
| 6 | Purchase merchandise | Accounts: 3400, 1576, 1600<br>Days: Monday - Sunday<br>Months: January – December 2023<br>Users: Bob, Charlie<br>Items: multiple<br>Non-working hours / weekend: 5%<br>Time delta: 90% | 2,000 |
| 6 | Payment | Accounts: 1600, 1100<br>rest as purchase merchandise | 2,000 |
| 7 | Purchase merchandise | Accounts: 3400, 1576, 1600<br>Days: Monday - Sunday<br>Months: January – December 2023<br>Users: Bob, Charlie<br>Items: multiple<br>Non-working hours / weekend: 5%<br>Time delta: 90% | 1,500 |
| 7 | Payment | Accounts: 1600, 1000<br>rest as purchase merchandise | 1,500 |
| 8 | Sales of goods | Accounts: 1400, 8400, 1776<br>Days: Monday - Sunday<br>Months: January – December 2023<br>Users: Bob, Charlie<br>Items: multiple<br>Non-working hours / weekend: 5%<br>Time delta: 90% | 1,000 |
| 8 | Incoming payment | Accounts: 1100, 1400<br>rest as sales of goods | 1,000 |
| 9 | Sales of goods | Accounts: 1400, 8400, 1776<br>Days: Monday - Friday<br>Months: January – December 2023<br>Users: Bob, Charlie<br>Items: single<br>Time delta: 80% | 500 |
| 9 | Incoming payment | Accounts: 1000, 1400<br>Days: Monday - Friday<br>Months: January – December 2023<br>Users: Bob, Charlie<br>Items: single<br>Non-working hours / weekend: 5%<br>Time delta: 80% | 500 |
| 10 | Different expenses | Accounts: 1000, 1200, 1576, 4100, 4120, 4130, 4210, 4240, 4250, 4260, 4360, 4580, 4600, 4710, 4930, 4950<br>Days: Monday - Friday<br>Months: January – December 2023<br>Users: Alice, Max<br>Items: single<br>Non-working hours: 48.7%<br>Time delta: 80% | 819 |

**TABLE 3**

**Evaluation results**

| Variant | Model | TP | FP | TN | FN | Precision | Recall | F1 | ROC-AUC |
|---|---|---|---|---|---|---|---|---|---|
| JET | - | 573 | 10,019 | 40,484 | 0 | - | - | - | - |
| ML | AE | 138 | 373 | 50,130 | 435 | **0.63** | 0.62 | 0.62 | 0.61 |
| | IF | 554 | 5,546 | 44,957 | 19 | 0.55 | **0.93** | 0.55 | 0.92 |
| | HS | 50 | 1,319 | 49,184 | 523 | 0.51 | 0.53 | 0.52 | 0.53 |
| H1 | AE | 138 | 55 | 50,448 | 435 | **0.85** | 0.62 | 0.68 | 0.61 |
| | IF | 554 | 1,214 | 49,289 | 19 | 0.66 | 0.97 | **0.73** | 0.97 |
| | HS | 50 | 123 | 50,380 | 523 | 0.64 | 0.54 | 0.56 | 0.54 |
| H2 | AE | 194 | 317 | 50,186 | 379 | 0.69 | 0.67 | 0.68 | 0.66 |
| | IF | 573 | 1,881 | 48,622 | 0 | 0.62 | **0.98** | 0.68 | **0.98** |
| | HS | 105 | 2,125 | 48,378 | 468 | 0.52 | 0.57 | 0.52 | 0.57 |